# PACGNet: Pyramidal Adaptive Cross-Gating Network for Multimodal Object Detection in Aerial Imagery

**Xvcheng Ning[1], Zidong Gu[2,*]**

[1] Department of Electronic Information, Handan Vocational College of Science and Technology, Handan 056046, China; 1092337597@qq.com

[2] School of Mathematics, China University of Mining and Technology, Xuzhou 221116, China; g445907022@gmail.com

[*] Correspondence: g445907022@gmail.com

**Abstract**

Object detection in aerial imagery is challenging because targets are often small, backgrounds are cluttered, and imaging conditions vary substantially. Multimodal sensing can improve robustness by providing complementary visual and spectral information, but effective fusion must balance cross-modal interaction with computational cost. This paper presents PACGNet, a lightweight backbone-centric framework for multimodal aerial object detection. Its Symmetrical Cross-Gating (SCG) module performs bidirectional gated interaction between the two modality streams at the same feature scale while retaining residual pathways for the original representations. The Pyramidal Feature-aware Multimodal Gating (PFMG) module subsequently forms a single fused feature pyramid by using the preceding higher-resolution level as a spatial prior for fusion at the next level. PACGNet achieves 82.2% mAP50 on the held-out DroneVehicle test set and 82.1% mAP50 on VEDAI with only 5.2 M parameters and 13.2 GFLOPs. These results indicate a favorable accuracy-efficiency trade-off for lightweight multimodal aerial object detection. Code is available at https://github.com/Alianess/PACGNet.

**Keywords:** multimodal object detection; RGB-infrared fusion; aerial imagery; cross-modality interaction; gating mechanisms

## 1. Introduction

Object detection is a fundamental computer-vision task with applications ranging from autonomous d riving [1] to unmanned aerial vehicle (UAV) reconnaissance [2]. Aerial imagery presents additional difficul ties because objects are often small [3], backgrounds are complex, and illumination varies considerably acr oss operating conditions [4]. Visible-light detectors provide texture and appearance cues [5], whereas infra red-only detectors can provide complementary information when visible-light cues are limited [6]. Multim odal fusion has therefore been widely studied for aerial object detection [7–9].

Existing multimodal detection methods can be broadly categorized as pixel-level [10,11], feature-level [12–14], and decision-level [15] fusion. Feature-level approaches are particularly common because they allo w information exchange at intermediate representation stages [16,17]. Many dual-stream architectures extr act features independently and then combine same-scale features by addition, concatenation, attention, or related interaction modules [7,18]. Two practical issues remain relevant for lightweight aerial detection. Fir st, unrestricted feature exchange can also transfer modality-specific interference or redundant responses [1 2,19]. Second, when fusion is performed independently at individual scales, relationships between adjacen t pyramid levels are not explicitly used during multimodal fusion, leaving most cross-level aggregation to the detection neck [20]. This motivates a fusion design that controls same-scale cross-modal interaction wh ile also exploiting the hierarchical structure of the backbone feature pyramid.

To address these issues, we propose the Pyramidal Adaptive Cross-Gating Network (PACGNet), whi ch performs multimodal fusion inside the backbone along two directions: cross-modal interaction at the sa me feature scale and hierarchical refinement across pyramid levels. The Symmetrical Cross-Gating (SCG) module is inserted at multiple scales of the dual-stream backbone and uses features from one modality to gate the enhancement of the other. Residual pathways retain the original modality representations while g ated branches regulate exchanged information. The Pyramidal Feature-aware Multimodal Gating (PFMG) module then completes the fusion before the neck by using the fused feature from the preceding higher-res olution level as a spatial prior for the current level. The resulting single-stream fused pyramid can be passed directly to a standard detection neck and head.

In summary, the contributions of this paper are as follows:

1. We propose PACGNet, a lightweight backbone-centric fusion framework that performs multimodal interaction within the backbone and outputs a single fused feature pyramid for a standard detection neck and head.

2. We design the SCG module for adaptive same-scale interaction between modality streams through bidirectional gating and residual feature preservation.

3. We develop the PFMG module to incorporate information from the preceding higher-resolution pyramid level when fusing features at the current level, providing explicit cross-level guidance during multimodal fusion.

4. Experiments on DroneVehicle and VEDAI show that PACGNet achieves 82.2% and 82.1% mAP50, respectively, with 5.2 M parameters and 13.2 GFLOPs, providing competitive detection accuracy under a lightweight computational budget.

The organization of the remaining sections of this paper is as follows: Section 2 reviews related work in multimodal fusion architectures, dynamic information filtering, and attention mechanisms. Section 3 provides a detailed description of our proposed PACGNet methodology, including its overall architecture, the Symmetrical Cross-Gating (SCG) module for horizontal feature interaction, and the Pyramidal Feature-aware Multimodal Gating (PFMG) module for vertical feature refinement. In Section 4, we present our experimental setup, including the datasets and evaluation metrics used, implementation details, and a comprehensive analysis of the results, along with an ablation study to validate the contributions of each component. Finally, Section 5 provides a conclusion summarizing the work and suggests directions for future research.

## 2. Related work

### *2.1. Multimodal Fusion Architectures for Object Detection*

RGB-Infrared (RGB-IR) fusion significantly enhances aerial object detection performance by combining the textural richness of visible light with the stability of thermal radiation. Early research primarily focused on feature-level fusion within dual-stream backbones. Qingyun et al. [18] proposed the Cross-modality Fusion Transformer (CFT), pioneering the use of self-attention to simultaneously model intra-modal long-range dependencies and inter-modal interactions for global context-aware fusion. Zhang et al. [8] introduced SuperYOLO, integrating a super-resolution branch into the YOLO framework to augment small-target features through infrared-visible fusion. Sun et al. [21] designed UA-CMDet, incorporating an uncertainty-aware module to quantify cross-modal detection confidence and suppress the influence of low-confidence targets. Transformer-based frameworks have emerged as a dominant trend due to their global modeling capabilities: Yuan and Wei [22] developed $C^2$Former, utilizing cross-modal attention and adaptive feature sampling for efficient feature alignment and fusion; Wang et al. [23] constructed FFODNet, which suppresses interfering features through joint expression optimization and task-specific enhancement modules. Recent studies emphasize lightweight design and hierarchical fusion: Bao et al. [12] proposed DDCINet, employing dual-dynamic cross-modal interaction to address modality inconsistency and redundancy; Liu et al. [24] designed a multi-branch progressive fusion network with a Modality Complementary Information Filter (MCIF) for adaptive fusion of modal advantages.

Recent methods have improved cross-modal alignment, attention, and feature selection, but same-scale interaction and cross-level fusion are often treated separately. PACGNet combines these two operations inside the backbone: SCG regulates same-scale feature exchange, while PFMG introduces hierarchical guidance across adjacent pyramid levels before the fused features enter the detection neck.

### *2.2. Dynamic Complementary Information Filtering and Noise Suppression*

Cross-modal fusion must retain useful complementary information without indiscriminately transferring redundant or modality-specific responses. Several approaches have been developed to regulate this information exchange:

Zhang et al. [25] proposed Guided Attentive Feature Fusion (GAFF), utilizing inter- and intra-modality attention modules to dynamically weigh and fuse multispectral features for pedestrian detection. Qingyun and Zhaokui [26] designed a Cross-Modal Attentive Feature Fusion (CMAFF) mechanism with differential enhancement and common feature selection modules. Recent approaches adopt learned alignment: Chen et al. [27] developed Offset-Aware Adaptive Feature Alignment (OAFA) to implicitly learn optimal fusion positions; Zhao et al. [7] leveraged visible-light reflectance features in RGFNet to guide cross-modal alignment; Ouyang et al. [28] addressed pretraining modality bias via cross-modal interaction in M²FP.

These studies motivate adaptive feature selection during multimodal interaction. SCG follows this direction with a lightweight bidirectional gating design: features from each stream provide guidance for the other stream, while residual connections retain the original representation. The module is intended to regulate cross-modal feature exchange without introducing a heavy attention block.

## 2.3. *Attention Mechanisms for Feature Optimization*

Attention mechanisms are widely used to reweight feature responses in complex visual scenes. Representative designs include the Convolutional Block Attention Module (CBAM) of Woo et al. [29], which combines channel and spatial attention, and ECA-Net of Wang et al. [30], which provides lightweight local cross-channel interaction.

To address the prevalence of small targets in aerial imagery—where features are easily overwhelmed by backgrounds—researchers have refined attention mechanisms: Zhang et al. [11] enhanced the Mamba architecture by integrating the Enhanced Small Target Detection (ESTD) module and Convolutional Attention Residual Gating (CARG) module to amplify small-target feature responses. Liu et al. [24] proposed Global-Local Synergistic Attention (GLSA), which models target-background contextual relationships via global-local fusion to improve feature discriminability. Lightweight designs further facilitate practical deployment: Zhang et al. [31] proposed the GHOST framework with hybrid quantization and knowledge distillation, preserving attention efficacy while ensuring efficiency.

Building on these ideas, PFMG extends feature weighting across pyramid levels rather than treating each scale independently. The fused feature from the preceding higher-resolution level is used as a spatial prior for fusion at the current level, providing hierarchical guidance as spatial resolution decreases.

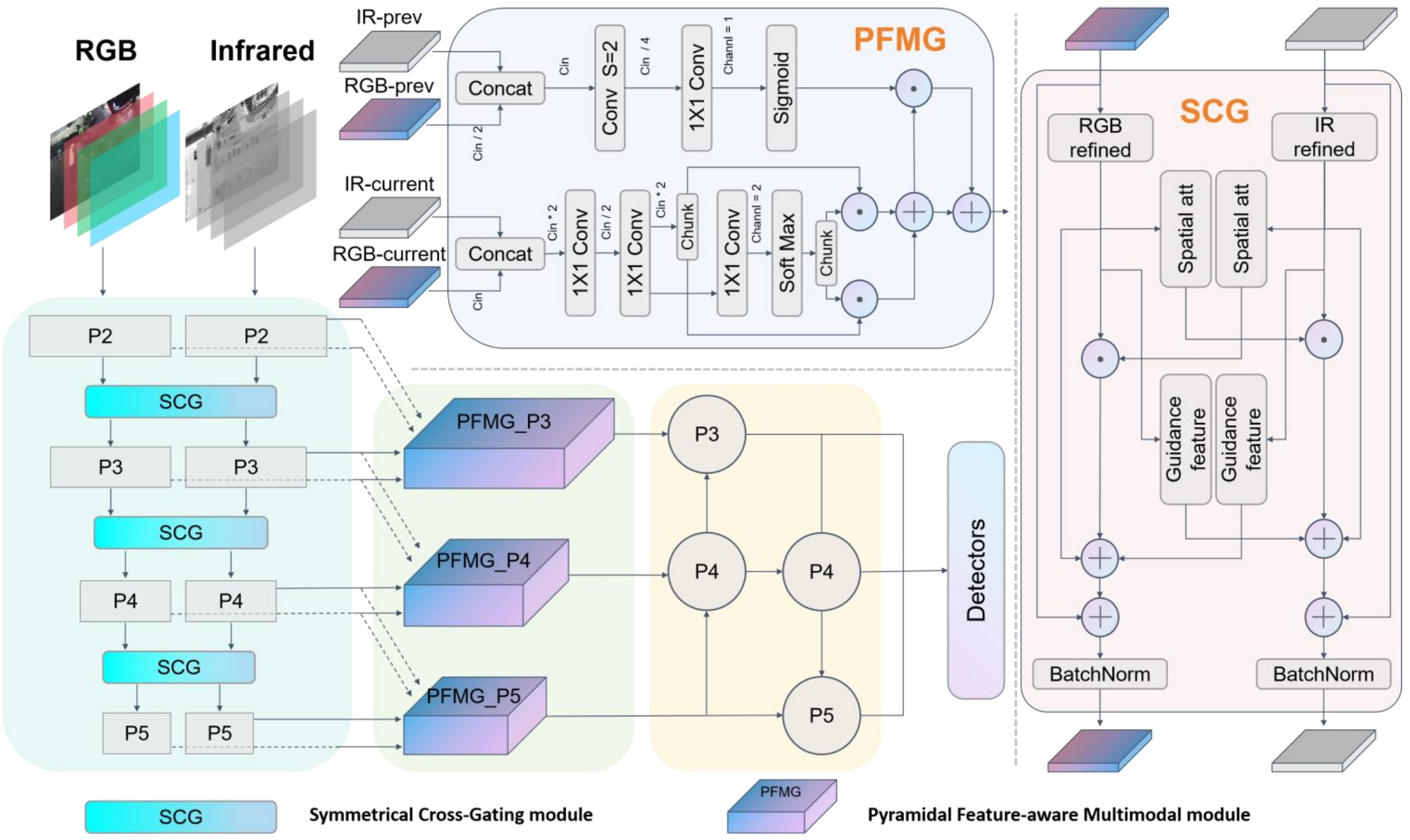

**Figure 1.** The overall framework of PACGNet. The network employs a dual-stream backbone to extract RGB and infrared features in parallel, interacting via Symmetrical Cross-Gating (SCG) modules. The SCG modules are shown schematically to represent the symmetrical, bi-directional enhancement between the two modalities. Subsequently, Pyramidal Feature-aware Multimodal Gating (PFMG) modules deeply fuse the multi-scale features, which are finally processed by a Path Aggregation Network (PAN) and fed to detectors for prediction.

## 3. Methodology

This section describes the architecture of the proposed Pyramidal Adaptive Cross-Gating Network (PACGNet). PACGNet organizes fusion in two directions: SCG performs same-scale cross-modal interaction, and PFMG performs cross-level hierarchical refinement. The overall architecture is introduced first, followed by the two modules and the loss function used for optimization.

### *3.1. Overall Architecture*

PACGNet uses a dual-stream backbone derived from YOLOv8 [32]. The two modality branches have the same topology but independent weights, and all parameters are initialized from scratch without external pre-training. As shown in Figure 1, each stream first extracts multi-scale features independently. SCG modules then regulate feature exchange between the two streams at multiple scales, after which PFMG modules form a single fused feature pyramid before the standard YOLOv8 neck and detection head.

The data flow is as follows:

3.1.1. Dual-Stream Backbone

The network begins with two parallel branches, each processing one modality (RGB or IR). Each branch consists of standard convolutional layers and C2f blocks to extract features at progressively coarser resolutions.

3.1.2. Inter-Modal Fusion with SCG

SCG modules are inserted after the P2, P3, and P4 feature stages. At each stage, features from the two modality streams provide bidirectional guidance to one another through gated interaction. This enables repeated same-scale information exchange during backbone feature extraction while retaining a residual path for each original stream.

3.1.3. Intra-Backbone Hierarchical Refinement with PFMG

After multi-scale features have been extracted, PFMG performs hierarchical fusion before the neck. The RGB and infrared features from the preceding higher-resolution level are jointly used to construct a spatial gate that guides fusion of the two modality features at the current lower-resolution level (for example, P3-level modality features guide P4 fusion). This process produces the fused feature maps P_fused_3, P_fused_4, and P_fused_5, which together form the single feature pyramid passed to the detection neck.

3.1.4. Detection Head

These highly-refined fused feature maps are subsequently passed to a standard YOLOv8 neck and detection head for final oriented object detection.

### *3.2. Symmetrical Cross-Gating (SCG) Module*

Before the final fusion, information must be exchanged between the two modality streams without simply passing all responses from one stream to the other. The Symmetrical Cross-Gating (SCG) module implements bidirectional feature interaction in which each modality generates guidance for the other. Figure 2 illustrates the IR-to-RGB path; the RGB-to-IR path is symmetric.

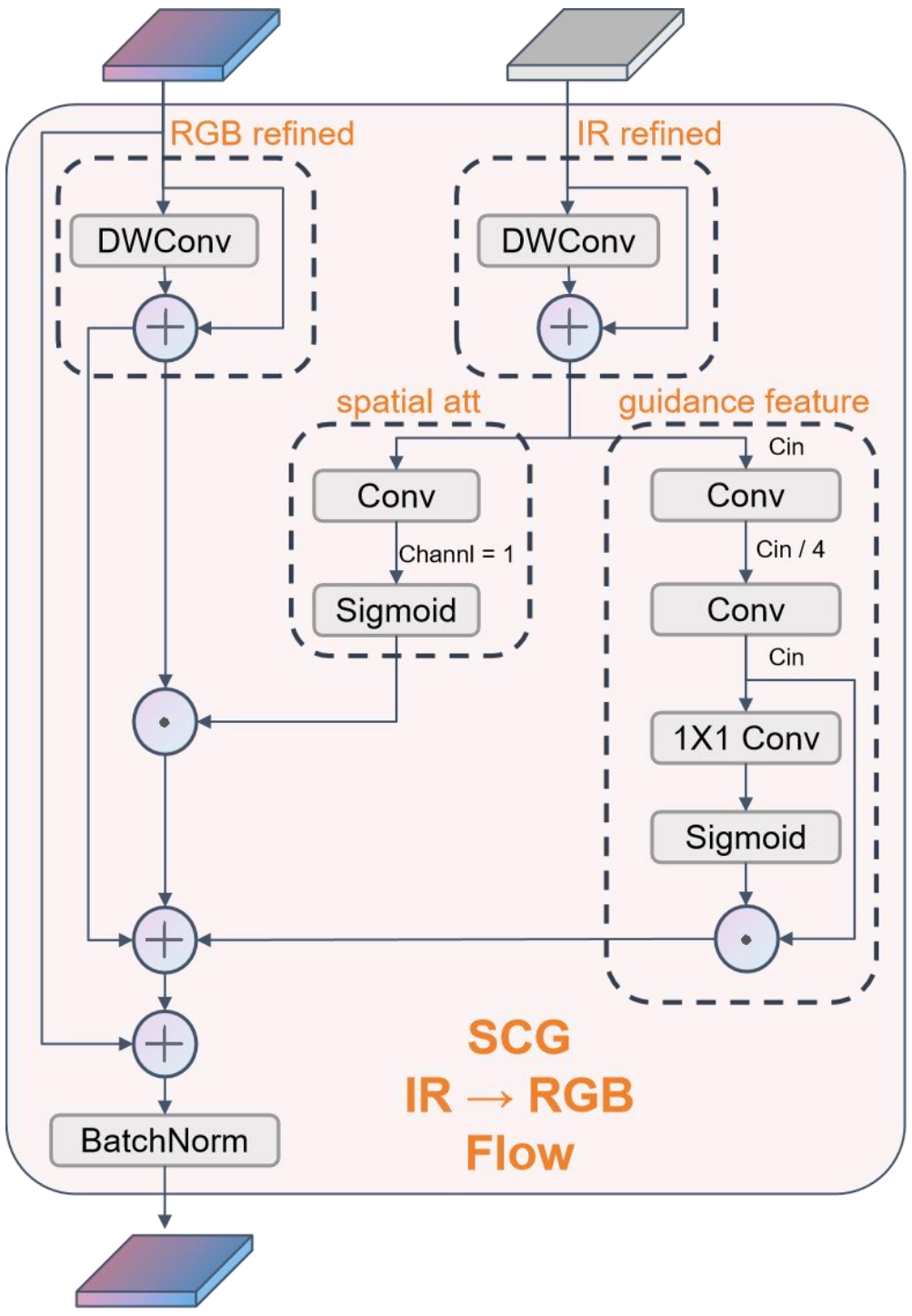

**Figure 2.** Detailed architecture of the Symmetrical Cross-Gating (SCG) module. Illustrating the IR-to-RGB guidance pa th, with its symmetrical counterpart omitted for brevity. The module utilizes a refined IR feature to generate parallel sp atial and channel-wise guidance. This guidance modulates and enhances the refined RGB feature, with a final residual connection preserving the original information.

**Step 1: Intra-Modal Feature Refinement:** Each input feature map, $F_{rgb}^{in}$ and $F_{ir}^{in}$, first passes through a Refined Feature Extractor. This extractor, composed of efficient Depthwise Separable Bottleneck blocks, enhances its intra-modal representation:

$$
\begin{aligned}
F'_{rgb} &= R(F_{rgb}^{in}) \\
F'_{ir} &= R(F_{ir}^{in})
\end{aligned}
\tag{1}
$$

where R(·) denotes the Refined Feature Extractor.

**Step 2: Cross-Modal Spatial Guidance:** A spatial attention map $M_S^{ir\to rgb}$ is generated from the refine d IR features to spatially modulate the refined RGB features. This map is generated by a 1x1 convolution fo llowed by a Sigmoid function. The modulation uses a (1 + M) scheme to ensure a residual connection, prev enting information loss if the gate is zero.

$$
\begin{aligned}
M_S^{ir\to rgb} &= \sigma(Conv_{1\times1}(F'_{ir})) \\
\tilde{F}_{rgb} &= F'_{rgb} \odot (1 + M_S^{ir\to rgb})
\end{aligned}
\tag{2}
$$

where σ is the Sigmoid function and ⊙ denotes element-wise multiplication.

**Step 3: Gated Cross-Modal Feature Guidance:** Concurrently, the refined IR feature is projected into a guidance feature through a lightweight projection block. A channel-wise gate is then generated from the p rojected guidance to regulate its contribution. The corresponding projection and gating operations are defi ned in Equation (3).

$$
\begin{aligned}
G_{rgb} &= P_{ir\to rgb}(F'_{ir}) \\
g_{rgb} &= \sigma(Conv_{1\times1}(G_{rgb}))
\end{aligned}
\tag{3}
$$

**Step 4: Final Fusion:** The spatially guided feature and gated cross-modal guidance are combined and added to the original RGB input through a residual connection. In Equation (4), Norm(·) denotes batch nor malization. The RGB-to-IR branch is defined symmetrically.

$$
F_{rgb}^{out} = Norm(F_{rgb}^{in} + (\tilde{F}_{rgb} + g_{rgb} \odot G_{rgb}))
\tag{4}
$$

### *3.3. Pyramidal Feature-aware Multimodal Gating (PFMG) Module*

A key design choice of PACGNet is to finalize multimodal fusion before the neck. The output is a single fused pyramid (P_fused_3, P_fused_4, P_fused_5), which allows the fusion backbone to be directly combined with any standard neck and head designed for single-modality inputs, such as the PANet of YOLOv8, without modifying the detection pipeline. This design improves the modularity and extensibility of the framework.

A common limitation of deep backbones is the progressive loss of spatial detail caused by repeated downsampling, which affects small objects most severely. To mitigate this, the PFMG module introduces a progressive, hierarchy-aware fusion strategy.

PFMG follows the natural hierarchy of the backbone, in which each pyramid level is generated from the preceding higher-resolution level. The paired RGB and infrared features at the preceding level are therefore used to construct spatial guidance for the fused representation at the current level. The notation for the RGB, IR, and fused features is given explicitly in Equations (5)–(7).

Step 1: Hierarchical Spatial Gate: The RGB and IR features from the preceding level are concatenated and passed through a Hierarchical Spatial Gate, implemented with a stride-2 convolution. This operation matches the spatial resolution of the current level and produces the spatial gate used in the current-level fusion, as defined in Equation (5).

$$M_S^{(i)} = H(Concat(F_{rgb}^{(i-1)}, F_{ir}^{(i-1)})) \quad (5)$$

where H(·) denotes the Hierarchical Spatial Gate.

Step 2: Modality Interaction and Adaptive Weighting: The current-level RGB and IR features are concatenated and processed by a modality-interaction block implemented with 1×1 convolutions. The output is split into interacted RGB and IR features. A 1×1 convolution followed by Softmax then produces pixel-wise modality weights that sum to one at each spatial location, as defined in Equation (6).

$$\begin{gathered} [F'_{rgb}, F'_{ir}] = Split(I(Concat(F_{rgb}^{(i)}, F_{ir}^{(i)}))) \\ [\omega_{rgb}^{(i)}, \omega_{ir}^{(i)}] = \mathrm{Soft\,max}(Conv_{1\times1}(Concat(F'_{rgb}, F'_{ir}))) \end{gathered} \quad (6)$$

**Step 3: Hierarchically Gated Fusion:** The fusion is a two-step process. First, a base fused feature is computed via a weighted average of the interacted features. Second, this representation is additively modulated by the hierarchical spatial gate from Step 1.

$$\begin{gathered} F_{base}^{(i)} = \omega_{rgb}^{(i)} \odot F'_{rgb} + \omega_{ir}^{(i)} \odot F'_{ir} \\ F_{fused}^{(i)} = F_{base}^{(i)} + M_S^{(i)} \odot F_{base}^{(i)} \end{gathered} \quad (7)$$

### *3.4. Loss Function*

For the models trained in this study (the dual-stream baseline, the ablation variants, and PACGNet), Wise-IoU (WIoU) v3 [33] is used in the bounding-box regression component. WIoU v3 employs a dynamic non-monotonic focusing mechanism based on the outlier degree of anchor boxes. The remaining detection-loss components, including the orientation-related prediction terms used by the YOLOv8-derived OBB implementation, are kept unchanged across these variants. Thus, the loss configuration is controlled consistently within our own experiments; loss settings of methods whose results are quoted from the literature are not assumed to be identical.

## 4. Experiments

This section elaborates on the experimental configuration, including the datasets employed and evaluation metrics. Subsequently, we present a comprehensive quantitative analysis, encompassing comparative assessments against relevant detectors, ablation studies investigating component contributions, and evaluations of model generalizability. Visualized results are also provided to offer intuitive insights into our findings.

### 4.1. Datasets and Evaluation Metrics

DroneVehicle. Our primary experiments use the DroneVehicle dataset [21], a public UAV visible-infrared vehicle-detection benchmark containing 28,439 registered RGB-infrared image pairs from diverse daytime and nighttime scenes and five vehicle categories. The dataset provides oriented bounding-box (OBB) annotations. Following the standard split, 17,990 pairs are used for training, 1,469 for validation, and 8,980 for testing. The validation split is used during model development, whereas the final PACGNet results reported in this paper are evaluated on the held-out 8,980-pair test set. The original 840 × 712 images are cropped to 640 × 512 effective regions and then resized to 640 × 640 for network input. Infrared annotations are used as the ground truth.

VEDAI. We further evaluate PACGNet on VEDAI [34], a multispectral aerial vehicle-detection benchmark containing visible and near-infrared (NIR) imagery at 1024 × 1024 resolution. After the standard filtering used in related VEDAI detection studies, 1,210 image pairs are evaluated with the benchmark's 10-fold protocol; each fold uses 1,089 images for training and 121 for testing. The 'plane' category is excluded from the reported eight-category mAP because of its extremely small sample count. We use the native 1024 × 1024 input resolution. The values reported for PACGNet in Table 2 follow this eight-category VEDAI setting.

**Evaluation Metrics**. For overall performance comparison, we adopt mean Average Precision at an IoU threshold of 0.5 (mAP50) as our primary evaluation metric. Additionally, we provide per-category AP50 scores to facilitate a finer-grained analysis of the model's performance on specific object classes. Specifically, mAP50 denotes the mean Average Precision calculated across all object categories at a fixed IoU threshold of 0.50, which determines true positives (TPs) and false positives (FPs), as we aim to focus on the most widely used benchmark in related multispectral object detection research.

### 4.2. Implementation Details

The proposed PACGNet model was implemented using the Ultralytics YOLOv8 v8.2.50 framework [32]. All experiments were executed on a server equipped with 8 NVIDIA GeForce RTX 3090 GPUs (24GB VRAM each), operating within a Python 3.10.15 environment. The model architecture builds upon the standard YOLOv8 dual-stream backbone, integrating our proposed PFMG (Pyramidal Feature-aware Multimodal Gating) and SCG (Symmetrical Cross-Gating) modules to effectively enhance cross-modal feature interaction and adaptive fusion. The archived DroneVehicle evaluation script uses the held-out test split with imgsz=640, batch=8, and IoU=0.5.

For the DroneVehicle experiments, model optimization was performed using the Stochastic Gradient Descent (SGD) optimizer with a batch size of 128 and 4 parallel data loading workers per GPU. Key hyperparameters included an initial learning rate (lr0) of 0.01, a final learning rate factor (lrf) of 0.01, momentum set to 0.937, and weight decay configured to 0.0005. A learning rate warmup schedule spanning 3.0 epochs was employed, featuring an initial momentum of 0.8 and an initial bias learning rate of 0.1. Training was conducted for a maximum of 300 epochs. For the VEDAI experiments, owing to the smaller scale of the dataset, models were trained on a single GPU with a reduced batch size of 8 at an input resolution of 1024 × 1024. Data augmentation strategies included Mosaic image composition, random horizontal and vertical flipping, and random translation.

### 4.3. Results Analysis

Table 1. Comparative performance of PACGNet and representative single- and multi-modality object detectors on the DroneVehicle test set. Metrics include per-category AP50 and overall mAP50 at IoU = 0.5. Best results among the listed methods are highlighted in red; second-best results are underlined.

| Detectors | Visible | Infrared | car | truck | bus | van | freight car | mAP50 |
|---|---|---|---|---|---|---|---|---|
| $S^2$A-Net[35] | √ | | 79.9 | 50.0 | 82.8 | 37.5 | 36.2 | 57.3 |
| Oriented R-CNN[19] | √ | | 80.1 | 53.8 | 85.4 | 43.3 | 41.6 | 60.8 |
| RoI Transformer[36] | √ | | 61.6 | 55.1 | 85.5 | 44.8 | 42.3 | 61.6 |
| YOLOv8n[32] | √ | | 96.2 | 72.7 | 94.5 | 54.4 | 53.2 | 74.2 |
| $S^2$A-Net[35] | | √ | 89.7 | 51.0 | 89.0 | 44.0 | 50.2 | 64.8 |

| Detectors | Visible | Infrared | car | truck | bus | van | freight car | mAP50 |
|---|---|---|---|---|---|---|---|---|
| Oriented R-CNN[19] | | √ | 89.8 | 57.4 | 89.3 | 45.4 | 53.1 | 67.0 |
| RoI Transformer[36] | | √ | 90.1 | 60.4 | 89.7 | 52.2 | 58.9 | 70.3 |
| YOLOv8n[32] | | √ | 97.7 | 75.5 | 94.7 | 57.6 | 61.2 | 77.4 |
| UA-CMDet[21] | √ | √ | 87.5 | 60.7 | 87.1 | 38.0 | 46.8 | 64.0 |
| LF-MDet[37] | √ | √ | 82.2 | 73.6 | 86.6 | 57.0 | 59.6 | 71.8 |
| $C^2$Former[22] | √ | √ | 90.2 | 68.3 | 89.8 | 58.5 | 64.4 | 74.2 |
| CALNet[38] | √ | √ | 90.3 | 76.2 | 89.1 | 58.5 | 63.0 | 75.4 |
| DDCINet[12] | √ | √ | 91.0 | 78.9 | 90.7 | 65.5 | 66.1 | 78.4 |
| M2FP[28] | √ | √ | 95.7 | 76.2 | 92.1 | 64.7 | 64.7 | 78.7 |
| OAFA[27] | √ | √ | 90.3 | 76.8 | 90.3 | <u>**66.0**</u> | <u>**73.3**</u> | 79.4 |
| MGMF[39] | √ | √ | 91.4 | 78.5 | 91.1 | 69.4 | 70.1 | 80.3 |
| RGFNet[7] | √ | √ | <u>**98.4**</u> | <u>**81.1**</u> | <u>**95.8**</u> | 63.0 | 68.7 | <u>**81.4**</u> |
| ESM-YOLO+[41] | √ | √ | **97.1** | **67.9** | **94.4** | 64.5 | 45.9 | **74.0** |
| PACGNet(our) | √ | √ | **98.5** | **83.0** | **96.3** | 64.3 | 68.7 | **82.2** |

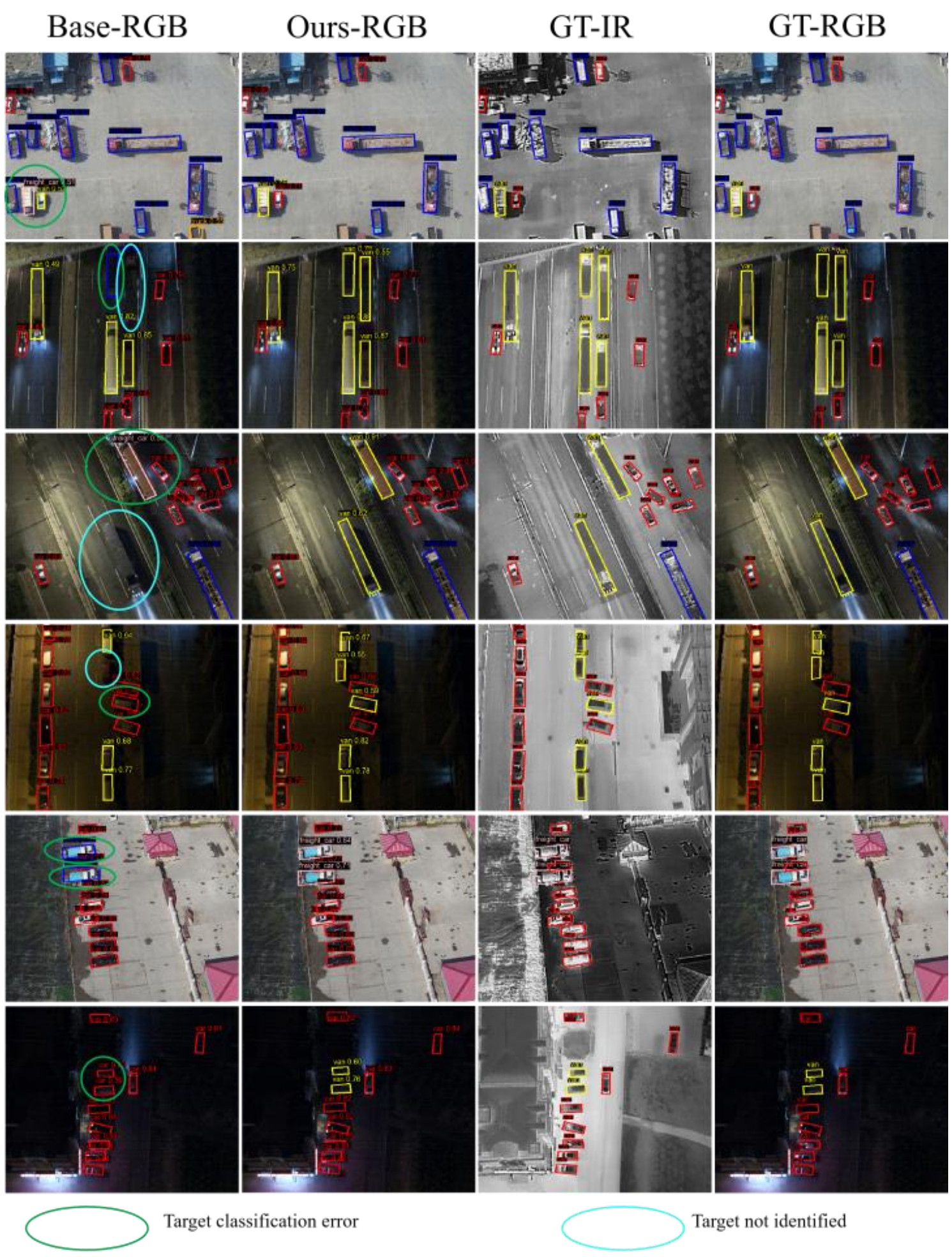

**Figure 3.** Visualization of the qualitative comparison between the baseline model and our improved PACGNet, alongs ide the ground truth, under various lighting conditions in the DroneVehicle dataset. The four columns, from left to rig ht, display: the detection results of the Baseline model, the detection results of the improved PACGNet, the ground tru th of the infrared image (GT-IR), and the ground truth of the visible light image (GT-RGB). The six rows illustrate six d ifferent lighting scenarios.

4.3.1 Performance on the DroneVehicle Dataset:

Table 1 reports results on the held-out DroneVehicle test set containing 8,980 image pairs. PACGNet r eaches 82.2% mAP50. Under our YOLOv8n single-modality baselines, the corresponding infrared-only res ult is 77.4% and the RGB-only result is 74.2%; PACGNet is therefore 4.8 percentage points above the infrar ed-only baseline and 8.0 points above the RGB-only baseline. Literature rows are reproduced from the cite d sources rather than recomputed from the displayed rounded category values.

Among the multimodal methods included in Table 1, PACGNet exceeds RGFNet (81.4%), MGMF (80. 3%), OAFA (79.4%), M2FP (78.7%), and DDCINet (78.4%). ESM-YOLO+ [41], a recent lightweight method with a parameter budget of approximately 5 M, reports 74.0% mAP50 in the comparison. These results pla ce PACGNet favorably among the listed methods while maintaining a small model size.

At the category level, PACGNet obtains 98.5% AP50 for car, 96.3% for bus, and 83.0% for truck. Perfor mance is lower for van (64.3%) and freight car (68.7%), where several comparison methods remain stronge r. This indicates that fine-grained discrimination between visually similar vehicle categories remains challe nging. Figure 3 provides qualitative examples across six lighting conditions and shows the corresponding baseline, PACGNet, and ground-truth results.

Table 2. Comparative performance of PACGNet and representative methods on the VEDAI dataset for oriented bound ing-box (OBB) detection. Metrics include per-category AP50 and overall mAP50. Best results among the listed methods are highlighted in red; second-best results are underlined.

| **Detectors** | **Visible** | **Infrared** | **car** | **truck** | **tractor** | **camping_car** | **van** | **vehicle** | **pickup** | **boat** | **mAP50** |
|---|---|---|---|---|---|---|---|---|---|---|---|
| YOLOv8n | √ | | 89.4 | 80.6 | 76.6 | 79.0 | **<u>83.0</u>** | 60.4 | 84.9 | 52.2 | 75.8 |
| YOLOv8n | | √ | 84.5 | 81.4 | 46.9 | 80.2 | **83.9** | 42.6 | 84.9 | 56.3 | 70.1 |
| YOLOFusion[26] | √ | √ | 91.7 | 78.1 | 71.9 | 78.9 | 75.2 | 54.7 | 85.9 | 71.7 | 75.9 |
| OST[31] | √ | √ | 91.1 | **<u>82.2</u>** | 84.9 | 74.9 | 82.9 | 64.6 | **87.7** | 60.2 | 78.6 |
| SuperYOLO[8] | √ | √ | 91.1 | 70.2 | 80.4 | 79.3 | 76.5 | 57.3 | 85.7 | 60.2 | 75.0 |
| ICAFusion[40] | √ | √ | - | - | - | - | - | - | - | - | 76.6 |
| $S_4^6$-MSTD[11] | √ | √ | **91.8** | 78.5 | **<u>85.8</u>** | **82.3** | 81.6 | **84.6** | 69.3 | **<u>75.6</u>** | **<u>81.2</u>** |
| MMYFNet[42] | √ | √ | - | **-** | **-** | **-** | - | **<u>-</u>** | **<u>-</u>** | **-** | **80.0** |
| PACGNet(our) | √ | √ | 90.5 | **87.1** | **87.4** | **80.8** | 82.5 | **<u>64.6</u>** | **<u>87.4</u>** | **76.8** | **82.1** |

4.3.2 Performance on the VEDAI Dataset:

Table 2 summarizes the VEDAI results under the eight-category setting. PACGNet obtains 82.1% mA P50. Among the literature results listed for comparison, S6_4-MSTD reports 81.2%, MMYFNet 80.0%, OST 78.6%, YOLOFusion 75.9%, and SuperYOLO 75.0%. Literature values are reproduced as reported by the cit ed sources; differences in implementation details and input settings should be considered when interpretin g cross-paper comparisons.

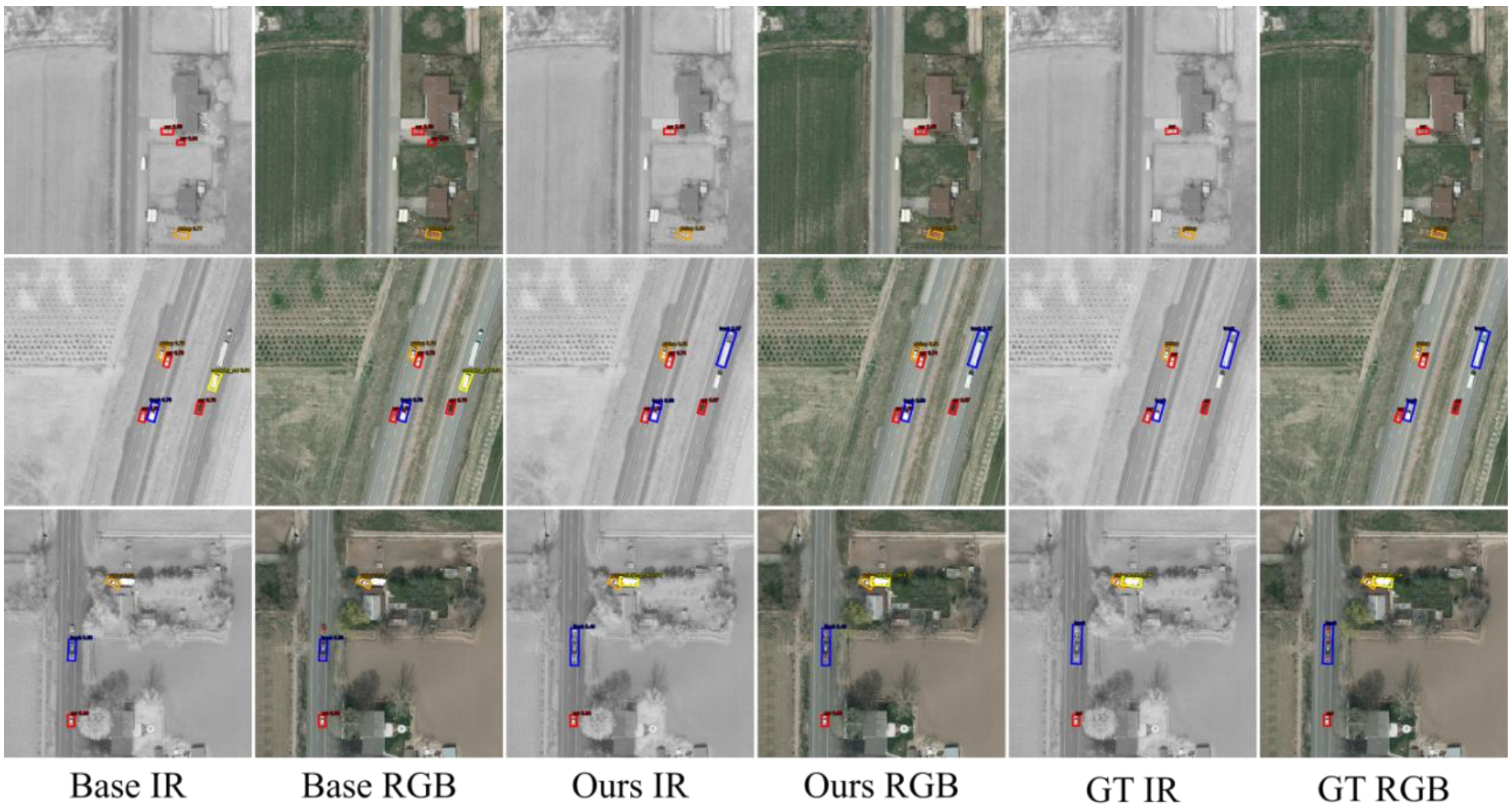

Figure 4. Qualitative comparison on the VEDAI dataset. The examples compare the baseline and PACGNet predictions with the corresponding ground truth, illustrating differences in missed detections, false positives, classification, and localization.

The category-level results provide additional detail. PACGNet obtains 87.1% AP50 for truck, 87.4% for tractor, and 76.8% for boat, while remaining close to the best listed results for pickup (87.4% versus 87.7%) and car (90.5% versus 91.8%). Because VEDAI provides visible and NIR imagery rather than thermal infrared imagery, these results are interpreted as evidence of complementary multispectral information rather than thermal contrast. The table also shows that performance varies across categories, with the generic vehicle class remaining comparatively difficult.

Table 3. Comparison of computational complexity and detection accuracy for PACGNet and selected multimodal detectors.

| Detectors | Params.(M) | GFLOPs | mAP50 |
|---|---|---|---|
| UA-CMDet[21] | 138.7 | - | 64.0 |
| $S^2$A-Net[35] | 38.6 | 93.0 | 64.8 |
| LF-MDet[37] | 38.7 | 77.7 | 71.8 |
| $C^2$Former[22] | 101.0 | 258.3 | 74.2 |
| MGMF[39] | 122.0 | - | 80.3 |
| PACGNet(our) | **5.2** | **13.2** | **82.2** |

Computational cost is important for deployment-oriented aerial detection. As shown in Table 3, PACGNet contains 5.2 M parameters and the archived 640 × 640 configuration is reported at 13.2 GFLOPs. This value is used as the model-complexity reference in this paper and should not be interpreted as the cost of inference at VEDAI's 1024 × 1024 input resolution. Cross-paper FLOP values are reproduced from the cited sources and may depend on implementation and profiling conventions. PACGNet nevertheless has a substantially smaller reported parameter count than C2Former (101.0 M) and MGMF (122.0 M), supporting the compactness of the proposed architecture.

### *4.4. Ablation Study*

To examine the contribution of the two proposed modules, PFMG and SCG are added individually and jointly to the same dual-stream YOLOv8 baseline. All variants use the same training and evaluation protocol on VEDAI and DroneVehicle. Table 4 reports mAP50 together with parameter count and computational cost.

**Table 4.** Ablation study for PFMG and SCG modules. Checkmarks (√) denote module inclusion. mAP(a): VEDAI; mAP (b): DroneVehicle. Red marks indicate the best performance.

| Method | PFMG | SCG | mAP(a) | mAP(b) | Params. | FLOPs |
|---|---|---|---|---|---|---|
| Baseline | | | 74.1 | 80.1 | 4.3M | **11.6G** |
| Baseline + PFMG | √ | | 76.7 | 80.8 | 4.7M | 12.3G |
| Baseline + SCG | | √ | 76.6 | 81.0 | 4.8M | 12.5G |
| PACGNet(our) | √ | √ | **82.1** | **82.2** | **5.2M** | 13.2G |

4.4.1. Effect of Individual Components:

The baseline obtains 74.1% mAP50 on VEDAI and 80.1% on DroneVehicle. The effects of adding each module individually are summarized below.

Baseline + PFMG: adding PFMG increases mAP50 to 76.7% on VEDAI (+2.6 points) and 80.8% on DroneVehicle (+0.7 points).

Baseline + SCG: adding SCG increases mAP50 to 76.6% on VEDAI (+2.5 points) and 81.0% on DroneVehicle (+0.9 points).

Both modules improve the baseline when introduced individually under the reported experimental setting. The larger gains on VEDAI are consistent with the motivation for using hierarchical and cross-modal feature interaction in scenes dominated by small targets, although the ablation results alone do not isolate the contribution by object size.

4.4.2. Combined Effect of PFMG and SCG:

When PFMG and SCG are used together, PACGNet reaches 82.1% mAP50 on VEDAI and 82.2% on DroneVehicle, corresponding to gains of 8.0 and 2.1 percentage points over the baseline, respectively.

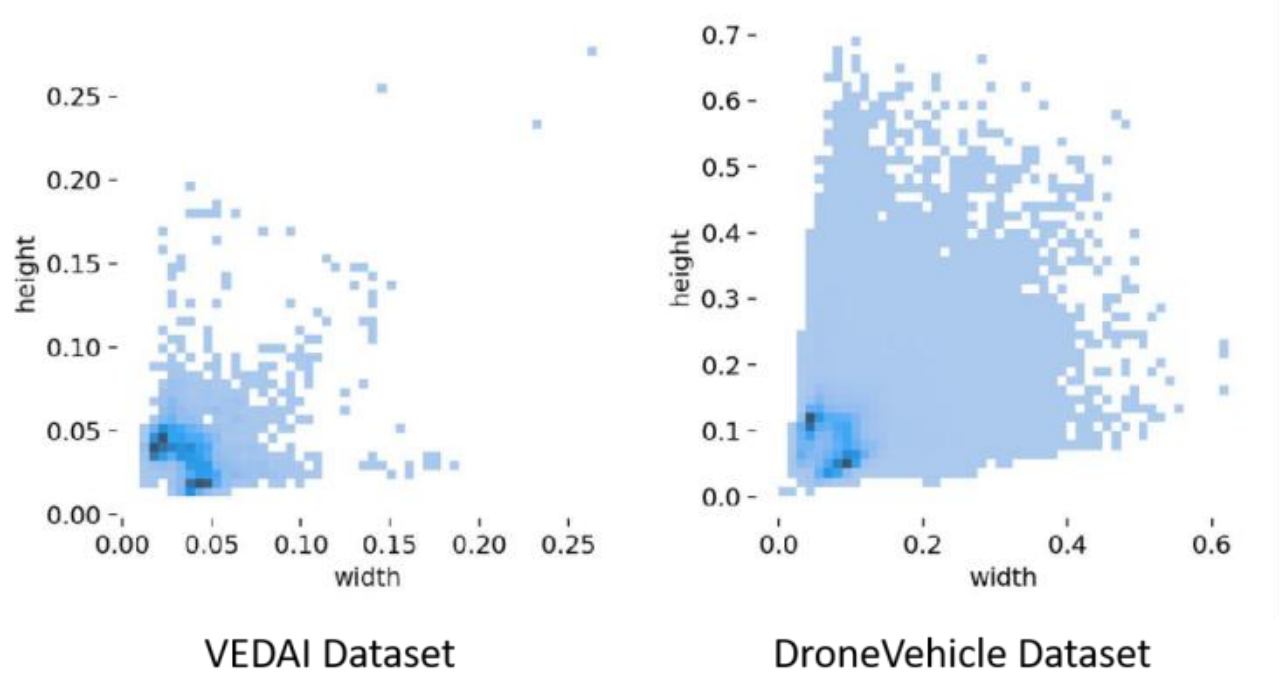

**Figure 5.** Comparison of ground truth bounding box size distributions between the DroneVehicle and VEDAI datasets.

The complete model provides the highest accuracy among the four ablation variants on both datasets. The improvement is particularly large on VEDAI, where many annotated objects occupy a small fraction of the image, as illustrated by the bounding-box size distribution in Figure 5. These results suggest that same-scale cross-modal interaction and cross-level hierarchical fusion can provide complementary functions within the proposed architecture. Statistical interaction between the two modules is not claimed from this single ablation setting.

4.4.3. Visualization Analysis:

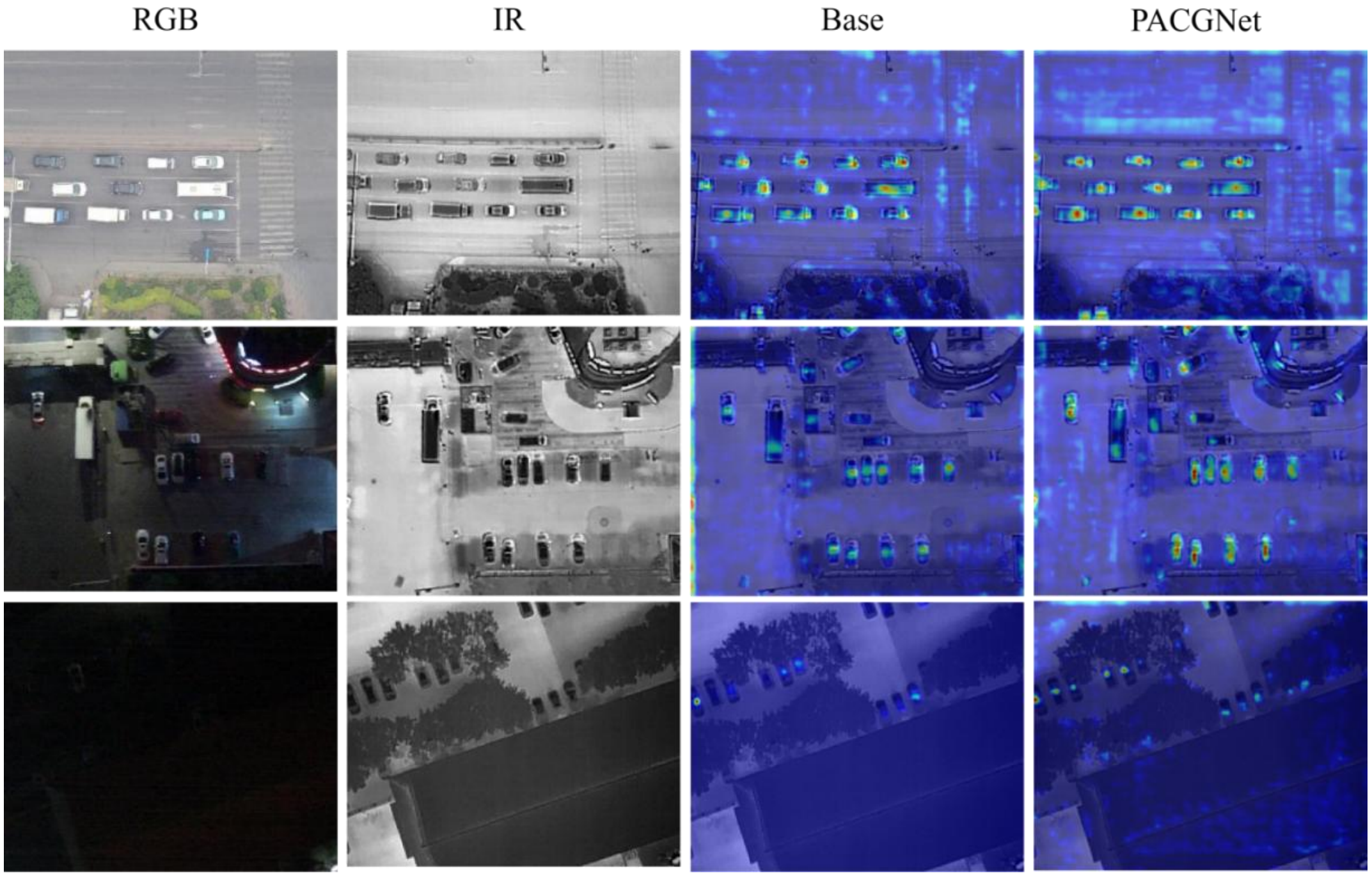

Figure 6. Feature-activation visualization for the baseline and PACGNet on identical input images. PACGNet shows more concentrated responses around vehicle regions in the illustrated examples.

Figure 6 compares feature-activation heatmaps from the baseline and PACGNet on identical inputs. In the illustrated examples, PACGNet produces more concentrated responses around vehicle regions and weaker activation in several background areas. This qualitative observation is consistent with the quantitative improvements reported in Tables 1, 2, and 4, but is not intended as a standalone causal analysis of the two modules.

## 5. Conclusions

This paper presented PACGNet, a lightweight backbone-centric network for multimodal object detection in aerial imagery. SCG performs gated same-scale interaction between the two modality streams, while PFMG constructs cross-level spatial guidance from the paired RGB and infrared features at the preceding higher-resolution level to guide fusion at the current level. Together, the two modules produce a single fused feature pyramid before the standard detection neck.

PACGNet achieves 82.2% mAP50 on the held-out DroneVehicle test set and 82.1% on VEDAI with 5.2M parameters and 13.2 GFLOPs. The experiments show competitive detection accuracy with a compact computational footprint, while the ablation study shows that both SCG and PFMG improve the corresponding dual-stream baseline under the reported setting. The method remains less effective for several fine-grained vehicle categories, and the current evaluation is limited to aerial vehicle-detection benchmarks. Future work will investigate stronger fine-grained feature discrimination and evaluate the fusion strategy on broader multispectral detection tasks.